\documentclass[10pt, a4paper]{article}
\usepackage{xcolor}
\usepackage{todonotes}
\usepackage{amsmath}
\usepackage{xurl}
\usepackage{amssymb}
\usepackage{times}
\usepackage{latexsym}
\usepackage{float}
\usepackage{pifont}
\usepackage{amssymb}
\usepackage[table]{xcolor}
\usepackage{enumitem}
\usepackage{multirow}
\usepackage{booktabs}
\usepackage{forest}
\usetikzlibrary{shadows}
\usepackage{times}
\usepackage{array}
\usepackage{latexsym}
\usepackage[utf8]{inputenc}
\usepackage[final]{lrec2026} 

\title{Counting on Consensus: Selecting the Right Inter-annotator Agreement Metric for NLP Annotation and Evaluation}

\name{Joseph James} 

\address{
  Department of Computer Science, The University of Sheffield \\
  Sheffield, United Kingdom \\
  \texttt{jhfjames1@sheffield.ac.uk}
}

\abstract{
Human annotation remains the foundation of reliable and interpretable data in Natural Language Processing (NLP). As annotation and evaluation tasks continue to expand, from categorical labelling to segmentation, subjective judgment, and continuous rating, measuring agreement between annotators has become increasingly more complex. This paper outlines how inter-annotator agreement (IAA) has been conceptualised and applied across NLP and related disciplines, describing the assumptions and limitations of common approaches. We organise agreement measures by task type and discuss how factors such as label imbalance and missing data influence reliability estimates. In addition, we highlight best practices for clear and transparent reporting, including the use of confidence intervals and the analysis of disagreement patterns. The paper aims to serve as a guide for selecting and interpreting agreement measures, promoting more consistent and reproducible human annotation and evaluation in NLP.
\\ \newline 
}

\begin{document}

\maketitleabstract

\section{Introduction}

Creating high-quality annotated data and ensuring reliable human evaluation are central to NLP, and the consistency of human judgments determines the validity of datasets and evaluations. IAA measures this consistency by quantifying the extent to which multiple annotators or evaluators apply the same labels to the same items. High IAA suggests clear guidelines and a reproducible results from texts or model outputs to human judgments, whereas low IAA can indicate under-specified instructions, insufficient training, or subjectivity in the task being evaluated \cite{hallgren2012computing}. However, relying on raw agreement alone can overestimate reliability, motivating the use of chance-corrected and task-appropriate statistics \cite{hayes2007answering,kohen1960coefficient}. The diversity of NLP tasks, from simple categorical labelling to span-based extraction, segmentation, and pairwise preferences, makes selecting the right agreement metric challenging. Recent work has highlighted how metric choice and distance functions affect interpretability and robustness for complex annotation tasks \cite{braylan2022measuring}, and evaluation studies document substantial variability in human criteria and correlations with automatic metrics \cite{chhun-etal-2022-human}. This highlights the need to match the agreement coefficient to the annotation design and to clarify what construct the metric is intended to capture  \cite{chhun-etal-2022-human}.

Another concern is the reporting practices. Point estimates without uncertainty overstate precision and reduce comparability. Methodological guidance recommends reporting confidence intervals, describing the rater design, and accounting for missing data and label prevalence, all of which affect interpretation of the results \cite{hallgren2012computing,hayes2007answering}. These considerations extend to human evaluation of model outputs, where pairwise preference studies reveal inconsistency under common setups and call for better design and meta-evaluation \cite{ghosh-etal-2024-compare}. Similar work shows that evaluator biases and inconsistencies can influence conclusions about system quality, meaning careful design is required to obtain reliable human judgments \cite{bavaresco-etal-2025-llms,liu2024aligning,humeval-2024-human,tam2024framework}. Finally, disagreement is not merely “noise”, modelling annotator or evaluator identities and perspectives can improve downstream learning and preserve legitimate variation in judgments, rather than collapsing it during aggregation \cite{deng-etal-2023-annotate}.

This paper offers an overview of how IAA can be measured in NLP. It reviews commonly used approaches across different settings, describing how their underlying assumptions affect interpretation and comparability. In addition, it outlines methodological considerations for reporting, including how to account for data imbalance, missing annotations, and uncertainty in agreement estimates. The aim is to help researchers select measures that align with their task and interpret them appropriately in context. By treating agreement not as an afterthought but as a key component, this work seeks to support more transparent and reproducible practices in NLP.

\section{Categorical data}
For tasks where each item is assigned a category, agreement can be measured in two main ways: by directly calculating the proportion of matching labels between annotators, or by using chance-corrected coefficients that adjust for agreement expected by random chance.

\subsection{Percentage Agreement}
Percentage agreement ($P_o$) is the simplest and most direct measure of inter-annotator reliability. It represents the proportion of items for which annotators assign the same label and is defined as the number of agreed items divided by the total number of items. However, because it does not account for agreement that could occur by chance, $P_o$ can overestimate reliability, especially when the categories are imbalanced. Despite this, it remains a useful baseline metric, particularly in exploratory or crowdsourced annotation studies. This measure provides an intuitive view of overall consistency and is often reported alongside chance-corrected statistics \cite{mohammad-etal-2018-semeval}.

\subsection{Bennett, Alpert, and Goldstein’s S}
Bennett, Alpert, and Goldstein’s $S$ coefficient provides a simple chance-corrected alternative to raw percentage agreement \cite{bennett}. It assumes that all categories are equally likely and is defined as
\[
S = 1 - \frac{k(1 - P_o)}{k - 1},
\]
where $k$ is the number of possible categories. The $S$ statistic adjusts for the probability of random agreement under the assumption of uniform category prevalence. However, it does not account for differences in annotator bias or imbalanced label distributions, which limits its suitability. Although rarely used in research, it is historically significant as one of the earliest attempts to correct for chance agreement metrics.

\subsection{Cohen’s Kappa}
Cohen’s $\kappa$ \cite{Cohen1960} is one of the earliest and most commonly used IAA measures for categorisation tasks. It applies when two annotators label the same set of items. $\kappa$ compares the observed agreement $P_o$ between the two annotators to the agreement expected by chance $P_e$, and is defined as
\[
\kappa = \frac{P_o - P_e}{1 - P_e},
\]
where \(P_o\) is the proportion of items the annotators agreed on, and \(P_e\) is the proportion expected to agree by random chance (based on each annotator’s labelling distribution). By subtracting \(P_e\), $\kappa$ corrects for random guessing. For example, if both annotators heavily favour one category, they might agree often simply by coincidence; $\kappa$ will reduce the score to account for this. $\kappa$ ranges from 1 (perfect agreement) down to 0 (no better than chance) or even negative (systematic disagreement). Cohen’s paper introduced $\kappa$ precisely to handle such situations where raw agreement is misleading.

\subsection{Fleiss’ Kappa}
Fleiss’ $\kappa$ \cite{Fleiss1971} generalises Cohen’s $\kappa$ to multiple annotators. It handles any number of annotators assigning nominal categories to a common set of items. Conceptually, Fleiss’ $\kappa$ also compares observed agreement to chance agreement, but it aggregates agreement across all annotators rather than comparing them pairwise. For each item \(i\), the observed agreement among \(N\) annotators is computed as

\[
P_i = \frac{1}{N(N-1)} \sum_{j=1}^{k} n_{ij}(n_{ij} - 1),
\]

where \(n_{ij}\) is the number of annotators who assigned item \(i\) to category \(j\). The mean observed agreement across all \(N\) items is then

\[
\bar{P} = \frac{1}{N} \sum_{i=1}^{N} P_i.
\]

The expected agreement \(P_e\) is derived from the overall category proportions, and the final $\kappa$ is computed in the same way as Cohen’s equation. Fleiss’ $\kappa$ assumes that each item receives the same number of annotations and can be sensitive to class imbalance or uneven marginal distributions. It is equivalent to Scott’s $\pi$ \cite{scott1955reliability} under the condition that each item has the same number of annotations.

\subsection{Weighted Kappa}
Weighted kappa \cite{cohen1968weighted} extends Cohen’s $\kappa$ to ordinal scales by assigning partial credit to near agreements. Instead of treating all disagreements equally, it applies a weighting function that penalises larger differences more heavily. The two most common weighting schemes are \textit{linear}, where weights decrease proportionally with distance, and \textit{quadratic}, where distant disagreements are penalised more strongly. Weighted $\kappa$ is particularly useful for rating tasks on Likert or ordinal scales, as it captures both the direction and magnitude of disagreement between annotators.

\subsection{Krippendorff’s Alpha}
Krippendorff’s $\alpha$ \cite{Krippendorff2013} is a versatile agreement metric that works for various data types (nominal, ordinal, interval, etc.), any number of annotators, and can handle missing data. $\alpha$ is popular in content analysis and increasingly in NLP because of this flexibility. It calculates agreement by focusing on disagreements:
\[
\alpha = 1 - \frac{D_o}{D_e},
\]
where $D_o$ is the observed disagreement and $D_e$ is the expected disagreement under chance. For nominal data, disagreement is usually defined as 0 if two labels are the same and 1 if different; for ordinal or interval data, a different distance function (e.g., squared difference) can be used. A major advantage of $\alpha$ is handling missing data. Notably, for nominal two-annotator cases with no missing values, $\alpha$ is mathematically equivalent to Cohen’s $\kappa$; for multiple annotators it relates closely to Fleiss’ $\kappa$. Researchers have used Krippendorff’s $\alpha$ in NLP annotation of discourse or subjective content, where not all annotators label everything or where an ordinal scale needs to be handled appropriately.

\subsection{Gwet’s AC1/AC2}

Gwet’s AC1/AC2 \cite{gwet2001handbook} are more recent chance-corrected agreement measures proposed to address some limitations of $\kappa$ in situations of high agreement or highly skewed class distributions (often called the “$\kappa$ paradox” where $\kappa$ can be low despite high observed agreement, or vice versa). Gwet’s AC1 (for nominal) and AC2 (for ordinal, with weights) use an alternative formula for expected agreement that tends to be more stable when classes are imbalanced or when annotators have bias. Recent NLP studies with extreme class imbalance have reported AC1 alongside $\kappa$ to provide a more nuanced view of agreement  \cite{chhun-etal-2024-language}.

\section{Structured annotations}
Not all annotation tasks involve assigning a single label to an item. In many NLP tasks, annotators mark segments of text, which introduces alignment issues. Examples include named entity recognition (NER), where annotators highlight spans corresponding to entities, or text segmentation, where annotators divide a document into segments (e.g., topic boundaries or dialogue turns).

\subsection{Span-Based Annotations}
For span labelling tasks, a common approach is to treat one annotator’s annotations as a \emph{predicted} set of spans and another’s as the \emph{gold} set, then compute precision, recall, and F$_1$ score between them. This pairwise F$_1$ (or the Dice coefficient) measures overlap agreement, it rewards the annotators for each entity span they both identified exactly and penalises cases where one annotator found an entity that the other missed or where their span boundaries differ. In practice, F$_1$ is calculated on a per-span basis (exact matches).

\subsection{Text Segmentation}
When the task is to segment a text into contiguous units (e.g., dividing an article into sections), annotators are placing boundaries in text. Two popular metrics for segmentation agreement are:
\begin{itemize}[noitemsep]
  \item \(P_k\) \cite{beeferman1999statistical}: Slides a fixed-size window through the text and checks if the segmentations agree on whether there is a boundary between two points.
  \item WindowDiff \cite{pevzner2002critique}: Similar to \(P_k\), but it penalizes near-misses less harshly.
\end{itemize}
Producing a score between 0 and 1, where 0 indicates perfect agreement. If annotators’ segment boundaries are slightly shifted, WindowDiff is more forgiving than \(P_k\). 

These metrics have been widely used in discourse and dialogue segmentation studies. For example, \citet{scaiano-inkpen-2012-getting} applied them to evaluate segment boundaries in multi-party conversation, highlighting how boundary-based agreement can reveal both annotator consistency and the inherent ambiguity of conversational structure. Building on these approaches, \citet{fournier-inkpen-2012-segmentation} introduced Segmentation Similarity, a generalised framework that unifies and extends \(P_k\) and WindowDiff by accounting for partial matches and variable boundary tolerance. 
  
\subsection{Unitising Tasks}
In some tasks, annotators not only mark segments of text but also decide the number and positions of those segments, so-called unitising annotations. For example, in discourse analysis, annotators might break a conversation into discourse units and label each unit with a discourse function. 

One metric is the holistic gamma (\(\gamma\)) measure \cite{Mathet2015}, which is designed for complex unitising tasks. Gamma treats the problem as a combination of segmentation and categorisation by finding an optimal alignment between units marked by different annotators and computing chance-corrected agreement. It accounts for both positional discrepancies (differences in unit boundaries) and categorical discrepancies (differences in labels).

\subsection{Boundary Edit Distance}
Boundary Edit Distance \cite{fournier2013evaluating} is a generalised measure of segmentation agreement that quantifies the minimal number of edits required to transform one annotator’s segmentation into another’s. It accounts for insertions, deletions, and near misses of boundaries, producing a flexible score that can handle varying degrees of segmentation granularity. Compared with WindowDiff or \(\gamma\), Boundary Edit Distance is often considered a more robust alternative for tasks such as discourse or dialogue segmentation, where partial overlaps and fuzzy boundaries are common.

\section{Continuous data}
Certain tasks require continuous or interval-scale outputs rather than discrete categories. Examples include grading the fluency or coherence of text, or scoring emotional intensity on a numeric scale \cite{abercrombie2023consistency, wong-paritosh-2022-k}. In such cases, the goal is to assess how consistently annotators assign scores along a continuous scale, rather than whether they select the same label. Measures of association or reliability for continuous data estimate how much of the variance in ratings reflects systematic differences between items rather than random noise or rater bias.

\subsection{Intraclass Correlation Coefficient}
The Intraclass Correlation Coefficient (ICC) is the most widely used reliability measure for continuous tasks. It quantifies the proportion of total variance in ratings that is attributable to true differences between items, relative to variance introduced by differences among raters or random error. Although well established in psychology and medicine, ICC is less frequently reported in NLP, where continuous annotation is often treated as ordinal or categorical.

Several ICC variants exist, distinguished by their model assumptions and by whether they estimate the consistency of individual ratings or the generalisability of aggregated ratings across multiple annotators. Following the framework introduced by \citet{shrout1979intraclass} and expanded by \citet{mcgraw1996forming}, the most common variants are:

\begin{itemize}[noitemsep]
    \item \textbf{ICC(1,1)}: One-way random effect, single measurement. Used when each item is rated by a different random set of raters.
    \item \textbf{ICC(1,k)}: One-way random effect, average measurement. It reflects the reliability of the mean rating when each item is rated by \(k\) raters.
    \item \textbf{ICC(2,1)}: Two-way random effect, single measurement (absolute agreement). Here, both items and raters are considered random samples from larger populations.
    \item \textbf{ICC(2,k)}: Two-way random effect, average measurement. It measures the reliability of the average rating from \(k\) raters.
    \item \textbf{ICC(3,1)}: Two-way mixed effect, single measurement (consistency). The raters are fixed (i.e., the particular raters are the only ones of interest) and the focus is on consistency rather than absolute agreement.
    \item \textbf{ICC(3,k)}: Two-way mixed effect, average measurement. It reflects the reliability of the average of \(k\) fixed raters.
\end{itemize}

In the two-way models, the difference between ICC(2,*) and ICC(3,*) lies in the treatment of the raters: in ICC(2,*) the raters are assumed to be randomly selected from a larger population, while in ICC(3,*) the raters are considered the only raters of interest (i.e., fixed). The choice between a single measurement (e.g., ICC(2,1)) and an average measurement (e.g., ICC(2,k)) depends on whether the study is concerned with the reliability of an individual rater's score or the aggregated score across raters. Selecting the appropriate ICC variant is essential for accurately assessing the reliability of annotations.

\subsection{Cronbach’s Alpha}
Cronbach’s $\alpha$ \cite{cronbach1951coefficient} is a measure of internal consistency that quantifies how closely related a set of ratings or items are as a group. It is mathematically equivalent to certain forms of the Intraclass Correlation Coefficient (ICC), specifically when the same raters assess each item under a one-way random-effects model. In annotation contexts, a high $\alpha$ suggests that annotators are applying the scale consistently across items. However, like ICC, its interpretation depends on model assumptions about rater and item effects, and it does not distinguish between consistency and absolute agreement.

\subsection{Concordance Correlation Coefficient}
The Concordance Correlation Coefficient (CCC) \cite{lawrence1989concordance} assesses how well two sets of continuous ratings agree in both precision and accuracy. Unlike standard correlation coefficients, which measure only the strength of association, CCC also captures deviation from the identity line, that is, how close annotators’ scores are to perfect concordance. It combines Pearson’s correlation with a term penalising mean and scale differences between annotators. CCC has been widely used in affective computing and emotion intensity annotation, where it provides a more stringent criterion for agreement than correlation alone.

\subsection{Correlation-Based Measures}

Correlation coefficients assess the degree to which annotators produce similar rating patterns across items. They are useful for measuring consistency rather than absolute agreement.

For ordinal or ranked judgments, non-parametric correlation coefficients such as Spearman’s $\rho$ \cite{ca468a70-0be4-389a-b0b9-5dd1ff52b33f} and Kendall’s $\tau$ \cite{kendall1938new} evaluate whether annotators preserve the same ordering of items, without assuming equal intervals between ranks. These measures are commonly used in human–model correlation studies, system ranking tasks, and as proxies for IAA when judgments are relative rather than absolute.

For continuous scores, Pearson’s correlation coefficient ($r$) \cite{pearson1895vii} measures the strength and direction of the linear relationship between annotators’ ratings. However, correlation alone does not indicate absolute agreement: two annotators may be perfectly correlated yet systematically offset in their scores. Therefore, correlation-based metrics should be interpreted as indicators of association rather than reliability in the strict sense.

\section{Metric Selection and Interpretation}

Selecting an appropriate IAA metric depends on the data type, number of annotators, and whether chance correction or missing data handling is essential. Table~\ref{tab:iaa-comparison} summarises commonly used metrics, outlining their key properties and limitations to support informed metric selection and interpretation.

\begin{table*}[ht]
\centering
\footnotesize
\setlength{\tabcolsep}{3pt}
\renewcommand{\arraystretch}{1.5}

\begin{tabular}{@{}p{2.1cm}
                p{2.0cm}
                p{1.1cm}
                p{2.0cm}
                p{1.5cm}
                p{2.0cm}
                p{3.8cm}@{}}
\toprule
\textbf{Metric} & \textbf{Data Type} & \textbf{Missing Data?} & \textbf{Number of Annotators?} & \textbf{Chance-Corrected?} & \textbf{Sensitive to Imbalance?} & \textbf{Limitations} \\ 
\midrule
\rowcolor{gray!5}
\textbf{Percentage Agreement} & Nominal & -- & Pairwise & -- & High & Overestimates reliability. \\ 
\textbf{Bennett’s S} & Nominal & -- & Two or more & \checkmark & -- & Assumes uniform label distribution. \\
\rowcolor{gray!5}
\textbf{Cohen’s $\kappa$} & Nominal & -- & Two only & \checkmark & High & Unstable when class frequencies or rater biases differ. \\
\textbf{Fleiss’ $\kappa$} & Nominal & -- & Three or more & \checkmark & High & Requires equal numbers of ratings per item. \\
\rowcolor{gray!5}
\textbf{Krippendorff’s $\alpha$} & Nominal / Ordinal / Interval / Ratio & \checkmark & Two or more & \checkmark & Moderate & Complex computation; interpretation can vary by distance metric. \\
\textbf{Gwet’s AC1/AC2} & Nominal / Ordinal & \checkmark & Two or more & \checkmark & Low & Less common in literature; parameter selection affects results. \\
\rowcolor{gray!5}
\textbf{Weighted $\kappa$} & Ordinal & -- & Two only & \checkmark & High & Requires careful choice of weighting scheme. \\
\textbf{ICC} & Continuous / Interval & \checkmark & Two or more & \checkmark & Moderate & Sensitive to model choice and assumes normality. \\
\rowcolor{gray!5}
\textbf{Cronbach’s $\alpha$} & Continuous / Interval & \checkmark & Two or more & \checkmark & -- & Assumes unidimensionality; may overestimate reliability. \\
\textbf{CCC} & Continuous & -- & Two only & \checkmark & Low & Sensitive to outliers; conflates accuracy and precision. \\
\rowcolor{gray!5}
\textbf{F$_1$ / Dice} & Span-based / Structured & -- & Pairwise & -- & Task-dependent & Sensitive to boundary mismatches.\\
\textbf{P$_k$ / WindowDiff} & Segmentation & -- & Pairwise & -- & -- & Results dependent on window size. \\
\rowcolor{gray!5}
\textbf{Gamma ($\gamma$)} & Unitising / Structured & \checkmark & Two or more & \checkmark & Moderate & Computationally intensive; conceptually complex. \\
\textbf{Boundary Edit Distance} & Segmentation & -- & Pairwise & -- & Task-dependent & Robust to near-misses; interpretation depends on tolerance. \\
\bottomrule
\end{tabular}

\caption{Overview of IAA metrics with key properties and limitations.}
\label{tab:iaa-comparison}
\end{table*}

\subsection{Interpretation of Agreement Scores}

The qualitative interpretation of IAA scores varies across research domains and metric families. For $\kappa$-type coefficients such as Cohen’s, Fleiss’, and Krippendorff’s~$\alpha$, conventions often follow the scale introduced by \citet{landis1977measurement} and later refined by \citet{viera2005understanding} and \citet{mchugh2012interrater}. These frameworks associate numerical ranges with descriptive categories such as “poor,” “fair,” “moderate,” “substantial,” and “almost perfect,” though their thresholds are not universally accepted.

For reliability coefficients such as the ICC and Cronbach’s~$\alpha$, psychometric and medical research typically applies more conservative standards, often considering values above 0.75 to indicate strong reliability \cite{koo2016guideline}. Correlation-based measures, including Pearson’s~$r$, Spearman’s~$\rho$, and Kendall’s~$\tau$, follow broadly similar conventions, distinguishing weak, moderate, and strong associations without fixed numeric boundaries.

Structured and segmentation-based metrics such as F$_1$, Boundary Similarity, Edit Distance, WindowDiff, P$_k$, and $\gamma$ lack standard interpretive scales. Their scores depend heavily on task setup, annotation granularity, and tolerance for partial matches, making relative rather than absolute comparisons more informative.

Recent work has raised broader concerns about how IAA scores are interpreted and reported. \citet{wong-etal-2021-cross} argue that fixed interpretive thresholds are overly rigid for complex or subjective tasks and propose replication-based benchmarks in place of universal cutoffs. \citet{klie2024analyzing} similarly show that many dataset papers report kappa-type coefficients without accounting for class imbalance, sample size, or annotator expertise, and call for greater transparency in reporting. Other studies highlight structural limitations: \citet{stefanovitch-piskorski-2023-holistic} find that traditional metrics often fail in multilingual or cross-domain settings, while \citet{li-etal-2024-estimating} demonstrate that categorical interpretive conventions do not directly transfer to structured or sequence-labelling tasks, where agreement is inherently lower due to hierarchical or contextual complexity.

Building on these critiques, \citet{richie-etal-2022-inter} emphasise that IAA should not be treated as a hard performance ceiling for machine learning systems, recommending instead the weighting or calibration of annotators rather than assuming uniform reliability. \citet{wong-paritosh-2022-k} introduce the $k$-rater reliability framework to quantify agreement across aggregated annotations, capturing stability that single-rater metrics often underestimate. Expanding this perspective, \citet{klie2024analyzing} provide a meta-analysis of 591 NLP datasets, revealing inconsistent quality control and advocating for standardised, transparent reporting of annotation design and error estimation.

Each agreement metric is based on different underlying assumptions about how to measure consistency between annotators. Some account for chance agreement, while others measure overlap or correlation directly. Because of these differences, agreement scores are not always comparable across metrics and must be interpreted in relation to the task and data. Results can also be affected by factors such as class imbalance, the number of annotators, and annotation granularity. Agreement should therefore be understood as a context-dependent indicator of reliability rather than an absolute measure of annotation quality.

\subsection{Reliability vs. Validity}

Reporting confidence intervals alongside IAA metrics is essential, as they indicate the range within which the true value of a metric is likely to lie and quantify uncertainty in its estimation. Narrow intervals suggest high precision, while wider intervals reflect greater variability or instability \cite{mchugh2012interrater}. Confidence intervals also enable more meaningful comparisons of agreement across tasks or groups by showing whether apparent differences are statistically significant or could arise from sampling variability \cite{reichenheim2004confidence}.

Although both p-values and confidence intervals support statistical inference, they serve different roles. A p-value assesses whether the observed agreement differs significantly from chance but provides no information about precision or effect magnitude. In contrast, a confidence interval conveys the range of plausible values for the agreement estimate, directly expressing its uncertainty. In IAA reporting, p-values can indicate significance, but confidence intervals offer a clearer sense of how stable and reliable the estimate is.

Inter-annotator agreement demonstrates reliability, the consistency with which annotators apply a scheme, but it does not establish validity. As \citet{artstein2008inter} note, high IAA only confirms that annotators are consistent, not that they are measuring the intended construct correctly. High agreement can arise even from oversimplified or biased guidelines, while low agreement may reflect genuine ambiguity or interpretive diversity rather than poor data quality \citep{plank-2022-problem}.

Researchers should therefore report reliability measures alongside evidence of validity, such as examples of ambiguous cases or comparisons to external references. Treating reliability and validity as complementary dimensions moves the field beyond numeric agreement scores and toward more interpretable and grounded evaluation practices \citep{howcroft-etal-2020-twenty}. Achieving both requires clear task definitions, empirical grounding, and iterative refinement through pilot studies, adjudication, and expert feedback to distinguish methodological uncertainty from genuine subjectivity.

\subsection{The Role of Disagreement}
Disagreement among annotators is a common feature of both annotation and evaluation in NLP. Rather than treating all divergence as noise, it can reveal task ambiguity, underspecified guidelines, or variation in rater preferences. Prior work identifies multiple sources of disagreement, including genuine linguistic ambiguity, inconsistent criteria, and contextual or interface effects that influence decision-making \cite{reidsma2008reliability,fleisig2023majority,xu-etal-2024-leveraging}. These findings caution against interpreting low agreement as evidence of poor data quality.

Recent research increasingly views disagreement as informative rather than erroneous. Retaining label distributions or “soft labels” allows ambiguity to be represented explicitly, while measures of label dispersion indicate where annotators diverge \cite{rodriguez-barroso-etal-2024-federated,fleisig2023majority}. Rater-aware models extend this approach by learning annotator representations, disentangling systematic bias from item difficulty, and improving both aggregation and downstream robustness \cite{fleisig2023majority}. Together, these methods treat disagreement as a meaningful signal about data, tasks, and annotator populations.

Practically, researchers are encouraged to report both agreement coefficients and measures of label dispersion, to analyse systematic disagreement across classes or subgroups, and to document annotation design factors such as training, and interface settings \cite{xu-etal-2024-leveraging,sandri-etal-2023-dont}. In subjective or interpretive tasks, preserving disagreement may be preferable to enforcing a single \emph{ground truth}, as it better reflects population diversity and produces models that are more robust and representative \cite{rodriguez-barroso-etal-2024-federated,sandri-etal-2023-dont,wan2023everyone}.

\subsection{Effects of Pay and Time Pressure}
Payment and time constraints influence both the quality of annotations and the behaviour of annotators. Early work showed that non-expert annotations can be valuable after aggregation, but quality varies with incentives and task design \cite{snow-etal-2008-cheap}. Flat-rate payment schemes are common yet often inequitable, as completion times vary widely, leading to very low effective wages for slower or more careful annotators and incentivising speed over accuracy \cite{salminen2023fair}. Performance-based or bonus systems can improve outcomes by promoting accuracy-oriented behaviour, though their effectiveness depends on task difficulty and the clarity of evaluation criteria \cite{rogstadius2011assessment}. Ethical analyses note that focusing solely on pay neglects broader issues of autonomy and bias, emphasising the need for transparency and worker agency \cite{shmueli-etal-2021-beyond}. From a motivation perspective, excessive reliance on extrinsic rewards can displace intrinsic motives such as curiosity, though clear feedback and transparent incentive structures can help counteract this effect.

Time constraints have comparable consequences. Tight deadlines promote heuristic or superficial judgments and reduce exploration under uncertainty, potentially inflating agreement for the wrong reasons \cite{wu2022time}. Conversely, strict viewing-time limits reduce both performance and satisfaction, while extended sessions without breaks cause fatigue and increase variance in responses \cite{lim2024towards,schlicher2021flexible}. Annotation difficulty also affects completion time, making uniform per-item pay problematic when item complexity varies \cite{wei2018clinical}. In practice, payment and timing protocols should prioritise accuracy over throughput: establish fair hourly rates based on observed completion times, define transparent evaluation criteria, avoid excessive time caps, and adjust compensation for more complex items. These design principles, reported alongside agreement statistics, support both reproducibility and ethical evaluation \cite{shmueli-etal-2021-beyond}.


\subsection{Human–Model Comparison}
Large Language Models (LLMs) are now used not only as systems to be evaluated but also as evaluators, producing judgments of fluency, coherence, or factuality that were once the exclusive domain of human annotators. This shift challenges the long-held assumption that human agreement represents the upper bound of evaluation quality. Recent studies demonstrate that model-based evaluation can equal or even exceed human reliability on several text quality dimensions \citep{gilardi2023chatgpt}. As a result, human annotations can no longer serve as an unquestioned gold standard when models are expected to complement or surpass human performance.

Surveys have shown that the emergence of LLM-based evaluation has reshaped the methodological landscape of NLP \citep{li2024llms,gu2024survey,chang2024survey,laskar-etal-2024-systematic}. These reviews highlight that while models often display higher internal consistency than human raters, they also risk reproducing systematic biases or calibration errors. Conversely, human disagreement can reflect genuine ambiguity or contextual sensitivity that models fail to capture. Rather than replacing human judgment, model-based evaluators should be benchmarked against diverse human perspectives and assessed for alignment with multiple criteria rather than correlation with a single reference set. Finally, \citet{bojic2025comparing} compare human annotators and several LLMs across sentiment, political leaning, emotional intensity, and sarcasm detection, finding that models often match human reliability on structured tasks but still underperform on nuanced or affective judgments, highlighting the continued value of human evaluation. 

\subsection{Annotator Expertise and Domain Knowledge}
Annotator experience and subject-matter knowledge strongly affect both annotation quality and reliability. In knowledge-intensive domains such as legal text, domain experts are often essential, as they can apply specialised concepts consistently and resolve ambiguities that non-experts may misinterpret \cite{yang2019predicting}. Even among experts, however, variability persists, as shown in medical imaging and other high-stakes tasks, reflecting the inherent complexity of annotation in such settings \cite{yang2023assessing}. For general tasks such as sentiment analysis or entity tagging, aggregated non-expert labels can achieve high reliability \cite{zlabinger2020dexa,snow-etal-2008-cheap}. Training and calibration also play a key role: novice annotators typically introduce higher variance, but consistency improves through structured feedback, curated examples, and repeated exposure \cite{huang2020coda,zlabinger2020dexa}.

Expertise, however, extends beyond factual knowledge to include interpretive perspective. Homogeneous expert groups may achieve high agreement but risk amplifying shared biases and overlooking alternative viewpoints. Mixed-expertise or crowdsourced settings tend to be noisier but are valuable for subjective or contested phenomena such as offensiveness, subjectivity, or inference, where disagreement may reflect genuine diversity rather than error \cite{mehta-srikumar-2023-verifying,snow-etal-2008-cheap}. Accordingly, documenting annotator backgrounds, training procedures, and recruitment strategies is critical for contextualising agreement scores and assessing generalisability. Transparent reporting of such information supports both reproducibility and ethical accountability in annotation research \cite{wang2013perspectives}.

Cultural and linguistic background also shapes annotation behaviour. \citet{hershcovich-etal-2022-challenges} emphasise that NLP must account for cultural as well as linguistic variation, since interpretations of meaning and social norms differ across communities. \citet{lee-etal-2024-exploring-cross} report large cross-country differences in English hate-speech labels, and \citet{pang2023auditing} find similar variation across cultural groups in human-labelling tasks. Together, these studies show that annotation reflects cultural perspective as much as linguistic competence and highlight the importance of documenting annotator backgrounds and considering cross-cultural diversity in dataset design.


\section{Conclusion}
IAA is fundamental to ensuring the reliability and reproducibility of annotated data in NLP, but its value depends on more than just reporting a number. Reliable evaluation requires aligning the agreement measure with the task and rater design, clearly communicating underlying assumptions, and reporting uncertainty to reflect the limits of precision. Equally important is examining patterns of disagreement to understand whether they reflect ambiguity, bias, or inconsistency. Treating IAA as an integral part of the methodological process can lead to more transparent and interpretable annotation and evaluation practices.

\section{Limitations}
This paper aims to provide an accessible overview of IAA measures used across different types of NLP annotation and evaluation tasks. As such, it focuses on breadth rather than depth, outlining commonly applied metrics and their practical implications. Readers seeking detailed mathematical descriptions or domain-specific adaptations are encouraged to explore the primary literature cited throughout this paper. Practitioners are additionally advised to carefully plan their annotation task before committing to a specific metric.

While we have aimed to cover a wide range of IAA metrics, it is possible that certain tasks or niche metrics have been omitted. Areas such as multimodal annotation, interactive evaluation, and subjective phenomena like humour or creativity may require approaches beyond those discussed in this paper.

\section*{Acknowledgements}
Joseph James is supported by the Centre for Doctoral Training in Speech and Language Technologies (SLT) and their Applications funded by UK Research and Innovation [grant number EP/S023062/1].

\nocite{*}
\section{Bibliographical References}\label{sec:reference}

\bibliographystyle{lrec2026-natbib}
\bibliography{lrec2026-example}

@inproceedings{pang2023auditing,
  title={Auditing cross-cultural consistency of human-annotated labels for recommendation systems},
  author={Pang, Rock Yuren and Cenatempo, Jack and Graham, Franklyn and Kuehn, Bridgette and Whisenant, Maddy and Botchway, Portia and Stone Perez, Katie and Koenecke, Allison},
  booktitle={Proceedings of the 2023 ACM Conference on Fairness, Accountability, and Transparency},
  pages={1531--1552},
  year={2023}
}

@inproceedings{lee-etal-2024-exploring-cross,
    title = "Exploring Cross-Cultural Differences in {E}nglish Hate Speech Annotations: From Dataset Construction to Analysis",
    author = "Lee, Nayeon  and
      Jung, Chani  and
      Myung, Junho  and
      Jin, Jiho  and
      Camacho-Collados, Jose  and
      Kim, Juho  and
      Oh, Alice",
    editor = "Duh, Kevin  and
      Gomez, Helena  and
      Bethard, Steven",
    booktitle = "Proceedings of the 2024 Conference of the North American Chapter of the Association for Computational Linguistics: Human Language Technologies (Volume 1: Long Papers)",
    month = jun,
    year = "2024",
    address = "Mexico City, Mexico",
    publisher = "Association for Computational Linguistics",
    url = "https://aclanthology.org/2024.naacl-long.236/",
    doi = "10.18653/v1/2024.naacl-long.236",
    pages = "4205--4224"
}

@inproceedings{hershcovich-etal-2022-challenges,
    title = "Challenges and Strategies in Cross-Cultural {NLP}",
    author = "Hershcovich, Daniel  and
      Frank, Stella  and
      Lent, Heather  and
      de Lhoneux, Miryam  and
      Abdou, Mostafa  and
      Brandl, Stephanie  and
      Bugliarello, Emanuele  and
      Cabello Piqueras, Laura  and
      Chalkidis, Ilias  and
      Cui, Ruixiang  and
      Fierro, Constanza  and
      Margatina, Katerina  and
      Rust, Phillip  and
      S{\o}gaard, Anders",
    editor = "Muresan, Smaranda  and
      Nakov, Preslav  and
      Villavicencio, Aline",
    booktitle = "Proceedings of the 60th Annual Meeting of the Association for Computational Linguistics (Volume 1: Long Papers)",
    month = may,
    year = "2022",
    address = "Dublin, Ireland",
    publisher = "Association for Computational Linguistics",
    url = "https://aclanthology.org/2022.acl-long.482/",
    doi = "10.18653/v1/2022.acl-long.482",
    pages = "6997--7013"
}

@article{li2024llms,
  title={Llms-as-judges: a comprehensive survey on llm-based evaluation methods},
  author={Li, Haitao and Dong, Qian and Chen, Junjie and Su, Huixue and Zhou, Yujia and Ai, Qingyao and Ye, Ziyi and Liu, Yiqun},
  journal={arXiv preprint arXiv:2412.05579},
  year={2024}
}

@article{gu2024survey,
  title={A survey on llm-as-a-judge},
  author={Gu, Jiawei and Jiang, Xuhui and Shi, Zhichao and Tan, Hexiang and Zhai, Xuehao and Xu, Chengjin and Li, Wei and Shen, Yinghan and Ma, Shengjie and Liu, Honghao and others},
  journal={arXiv preprint arXiv:2411.15594},
  year={2024}
}

@article{gilardi2023chatgpt,
  title={ChatGPT outperforms crowd workers for text-annotation tasks},
  author={Gilardi, Fabrizio and Alizadeh, Meysam and Kubli, Ma{\"e}l},
  journal={Proceedings of the National Academy of Sciences},
  volume={120},
  number={30},
  pages={e2305016120},
  year={2023},
  publisher={National Academy of Sciences}
}

@inproceedings{laskar-etal-2024-systematic,
    title = "A Systematic Survey and Critical Review on Evaluating Large Language Models: Challenges, Limitations, and Recommendations",
    author = "Laskar, Md Tahmid Rahman  and
      Alqahtani, Sawsan  and
      Bari, M Saiful  and
      Rahman, Mizanur  and
      Khan, Mohammad Abdullah Matin  and
      Khan, Haidar  and
      Jahan, Israt  and
      Bhuiyan, Amran  and
      Tan, Chee Wei  and
      Parvez, Md Rizwan  and
      Hoque, Enamul  and
      Joty, Shafiq  and
      Huang, Jimmy",
    editor = "Al-Onaizan, Yaser  and
      Bansal, Mohit  and
      Chen, Yun-Nung",
    booktitle = "Proceedings of the 2024 Conference on Empirical Methods in Natural Language Processing",
    month = nov,
    year = "2024",
    address = "Miami, Florida, USA",
    publisher = "Association for Computational Linguistics",
    url = "https://aclanthology.org/2024.emnlp-main.764/",
    doi = "10.18653/v1/2024.emnlp-main.764",
    pages = "13785--13816"
}

@article{chang2024survey,
  title={A survey on evaluation of large language models},
  author={Chang, Yupeng and Wang, Xu and Wang, Jindong and Wu, Yuan and Yang, Linyi and Zhu, Kaijie and Chen, Hao and Yi, Xiaoyuan and Wang, Cunxiang and Wang, Yidong and others},
  journal={ACM transactions on intelligent systems and technology},
  volume={15},
  number={3},
  pages={1--45},
  year={2024},
  publisher={ACM New York, NY}
}

@article{scott1955reliability,
  title={Reliability of content analysis: The case of nominal scale coding},
  author={Scott, William A},
  journal={Public opinion quarterly},
  pages={321--325},
  year={1955},
  publisher={JSTOR}
}

@article{klie2024analyzing,
  title={Analyzing dataset annotation quality management in the wild},
  author={Klie, Jan-Christoph and Castilho, Richard Eckart de and Gurevych, Iryna},
  journal={Computational Linguistics},
  volume={50},
  number={3},
  pages={817--866},
  year={2024},
  publisher={MIT Press 255 Main Street, 9th Floor, Cambridge, Massachusetts 02142, USA~…}
}

@article{bojic2025comparing,
  title={Comparing large Language models and human annotators in latent content analysis of sentiment, political leaning, emotional intensity and sarcasm},
  author={Boji{\'c}, Ljubi{\v{s}}a and Zagovora, Olga and Zelenkauskaite, Asta and Vukovi{\'c}, Vuk and {\v{C}}abarkapa, Milan and Veseljevi{\'c} Jerkovi{\'c}, Selma and Jovan{\v{c}}evi{\'c}, Ana},
  journal={Scientific reports},
  volume={15},
  number={1},
  pages={11477},
  year={2025},
  publisher={Nature Publishing Group UK London}
}

@article{reichenheim2004confidence,
  title={Confidence intervals for the kappa statistic},
  author={Reichenheim, Michael E},
  journal={The Stata Journal},
  volume={4},
  number={4},
  pages={421--428},
  year={2004},
  publisher={SAGE Publications Sage CA: Los Angeles, CA}
}

@article{abercrombie2023consistency,
  title={Consistency is key: Disentangling label variation in natural language processing with intra-annotator agreement},
  author={Abercrombie, Gavin and Rieser, Verena and Hovy, Dirk},
  journal={arXiv preprint arXiv:2301.10684},
  year={2023}
}

@inproceedings{wong-paritosh-2022-k,
    title = "k-{R}ater {R}eliability: {T}he Correct Unit of Reliability for Aggregated Human Annotations",
    author = "Wong, Ka  and
      Paritosh, Praveen",
    editor = "Muresan, Smaranda  and
      Nakov, Preslav  and
      Villavicencio, Aline",
    booktitle = "Proceedings of the 60th Annual Meeting of the Association for Computational Linguistics (Volume 2: Short Papers)",
    month = may,
    year = "2022",
    address = "Dublin, Ireland",
    publisher = "Association for Computational Linguistics",
    url = "https://aclanthology.org/2022.acl-short.42/",
    doi = "10.18653/v1/2022.acl-short.42",
    pages = "378--384"
}

@inproceedings{richie-etal-2022-inter,
    title = "Inter-annotator agreement is not the ceiling of machine learning performance: Evidence from a comprehensive set of simulations",
    author = "Richie, Russell  and
      Grover, Sachin  and
      Tsui, Fuchiang (Rich)",
    editor = "Demner-Fushman, Dina  and
      Cohen, Kevin Bretonnel  and
      Ananiadou, Sophia  and
      Tsujii, Junichi",
    booktitle = "Proceedings of the 21st Workshop on Biomedical Language Processing",
    month = may,
    year = "2022",
    address = "Dublin, Ireland",
    publisher = "Association for Computational Linguistics",
    url = "https://aclanthology.org/2022.bionlp-1.26/",
    doi = "10.18653/v1/2022.bionlp-1.26",
    pages = "275--284"
}

@inproceedings{fournier-inkpen-2012-segmentation,
    title = "Segmentation Similarity and Agreement",
    author = "Fournier, Chris  and
      Inkpen, Diana",
    editor = "Fosler-Lussier, Eric  and
      Riloff, Ellen  and
      Bangalore, Srinivas",
    booktitle = "Proceedings of the 2012 Conference of the North {A}merican Chapter of the Association for Computational Linguistics: Human Language Technologies",
    month = jun,
    year = "2012",
    address = "Montr{\'e}al, Canada",
    publisher = "Association for Computational Linguistics",
    url = "https://aclanthology.org/N12-1016/",
    pages = "152--161"
}

@inproceedings{scaiano-inkpen-2012-getting,
    title = "Getting More from Segmentation Evaluation",
    author = "Scaiano, Martin  and
      Inkpen, Diana",
    editor = "Fosler-Lussier, Eric  and
      Riloff, Ellen  and
      Bangalore, Srinivas",
    booktitle = "Proceedings of the 2012 Conference of the North {A}merican Chapter of the Association for Computational Linguistics: Human Language Technologies",
    month = jun,
    year = "2012",
    address = "Montr{\'e}al, Canada",
    publisher = "Association for Computational Linguistics",
    url = "https://aclanthology.org/N12-1038/",
    pages = "362--366"
}

@inproceedings{stefanovitch-piskorski-2023-holistic,
    title = "Holistic Inter-Annotator Agreement and Corpus Coherence Estimation in a Large-scale Multilingual Annotation Campaign",
    author = "Stefanovitch, Nicolas  and
      Piskorski, Jakub",
    editor = "Bouamor, Houda  and
      Pino, Juan  and
      Bali, Kalika",
    booktitle = "Proceedings of the 2023 Conference on Empirical Methods in Natural Language Processing",
    month = dec,
    year = "2023",
    address = "Singapore",
    publisher = "Association for Computational Linguistics",
    url = "https://aclanthology.org/2023.emnlp-main.6/",
    doi = "10.18653/v1/2023.emnlp-main.6",
    pages = "71--86"
}

@inproceedings{li-etal-2024-estimating,
    title = "Estimating Agreement by Chance for Sequence Annotation",
    author = "Li, Diya  and
      Rose, Carolyn  and
      Yuan, Ao  and
      Zhou, Chunxiao",
    editor = "Ku, Lun-Wei  and
      Martins, Andre  and
      Srikumar, Vivek",
    booktitle = "Proceedings of the 62nd Annual Meeting of the Association for Computational Linguistics (Volume 1: Long Papers)",
    month = aug,
    year = "2024",
    address = "Bangkok, Thailand",
    publisher = "Association for Computational Linguistics",
    url = "https://aclanthology.org/2024.acl-long.278/",
    doi = "10.18653/v1/2024.acl-long.278",
    pages = "5085--5097"
}

@article{cohen1968weighted,
  title={Weighted kappa: Nominal scale agreement provision for scaled disagreement or partial credit.},
  author={Cohen, Jacob},
  journal={Psychological bulletin},
  volume={70},
  number={4},
  pages={213},
  year={1968},
  publisher={American Psychological Association}
}

@article{artstein2008inter,
  title={Inter-coder agreement for computational linguistics},
  author={Artstein, Ron and Poesio, Massimo},
  journal={Computational linguistics},
  volume={34},
  number={4},
  pages={555--596},
  year={2008},
  publisher={MIT Press One Rogers Street, Cambridge, MA 02142-1209, USA journals-info~…}
}

@inproceedings{howcroft-etal-2020-twenty,
    title = "Twenty Years of Confusion in Human Evaluation: {NLG} Needs Evaluation Sheets and Standardised Definitions",
    author = "Howcroft, David M.  and
      Belz, Anya  and
      Clinciu, Miruna-Adriana  and
      Gkatzia, Dimitra  and
      Hasan, Sadid A.  and
      Mahamood, Saad  and
      Mille, Simon  and
      van Miltenburg, Emiel  and
      Santhanam, Sashank  and
      Rieser, Verena",
    editor = "Davis, Brian  and
      Graham, Yvette  and
      Kelleher, John  and
      Sripada, Yaji",
    booktitle = "Proceedings of the 13th International Conference on Natural Language Generation",
    month = dec,
    year = "2020",
    address = "Dublin, Ireland",
    publisher = "Association for Computational Linguistics",
    url = "https://aclanthology.org/2020.inlg-1.23/",
    doi = "10.18653/v1/2020.inlg-1.23",
    pages = "169--182"
}

@inproceedings{plank-2022-problem,
    title = "The ``Problem'' of Human Label Variation: On Ground Truth in Data, Modeling and Evaluation",
    author = "Plank, Barbara",
    editor = "Goldberg, Yoav  and
      Kozareva, Zornitsa  and
      Zhang, Yue",
    booktitle = "Proceedings of the 2022 Conference on Empirical Methods in Natural Language Processing",
    month = dec,
    year = "2022",
    address = "Abu Dhabi, United Arab Emirates",
    publisher = "Association for Computational Linguistics",
    url = "https://aclanthology.org/2022.emnlp-main.731/",
    doi = "10.18653/v1/2022.emnlp-main.731",
    pages = "10671--10682"
}

@inproceedings{wong-etal-2021-cross,
    title = "Cross-replication Reliability - An Empirical Approach to Interpreting Inter-rater Reliability",
    author = "Wong, Ka  and
      Paritosh, Praveen  and
      Aroyo, Lora",
    editor = "Zong, Chengqing  and
      Xia, Fei  and
      Li, Wenjie  and
      Navigli, Roberto",
    booktitle = "Proceedings of the 59th Annual Meeting of the Association for Computational Linguistics and the 11th International Joint Conference on Natural Language Processing (Volume 1: Long Papers)",
    month = aug,
    year = "2021",
    address = "Online",
    publisher = "Association for Computational Linguistics",
    url = "https://aclanthology.org/2021.acl-long.548/",
    doi = "10.18653/v1/2021.acl-long.548",
    pages = "7053--7065"
}

@article{mchugh2012interrater,
  title={Interrater reliability: the kappa statistic},
  author={McHugh, Mary L},
  journal={Biochemia medica},
  volume={22},
  number={3},
  pages={276--282},
  year={2012},
  publisher={Hrvatsko dru{\v{s}}tvo za medicinsku biokemiju i laboratorijsku medicinu}
}

@article{viera2005understanding,
  title={Understanding interobserver agreement: the kappa statistic},
  author={Viera, Anthony J and Garrett, Joanne M and others},
  journal={Fam med},
  volume={37},
  number={5},
  pages={360--363},
  year={2005}
}

@article{landis1977measurement,
  title={The measurement of observer agreement for categorical data},
  author={Landis, J Richard and Koch, Gary G},
  journal={biometrics},
  pages={159--174},
  year={1977},
  publisher={JSTOR}
}

@article{mcgraw1996forming,
  title={Forming inferences about some intraclass correlation coefficients.},
  author={McGraw, Kenneth O and Wong, Seok P},
  journal={Psychological methods},
  volume={1},
  number={1},
  pages={30},
  year={1996},
  publisher={American Psychological Association}
}

@article{shrout1979intraclass,
  title={Intraclass correlations: uses in assessing rater reliability.},
  author={Shrout, Patrick E and Fleiss, Joseph L},
  journal={Psychological bulletin},
  volume={86},
  number={2},
  pages={420},
  year={1979},
  publisher={American Psychological Association}
}

@article{pearson1895vii,
  title={VII. Note on regression and inheritance in the case of two parents},
  author={Pearson, Karl},
  journal={proceedings of the royal society of London},
  volume={58},
  number={347-352},
  pages={240--242},
  year={1895},
  publisher={The Royal Society London}
}

@article{kendall1938new,
  title={A new measure of rank correlation},
  author={Kendall, Maurice G},
  journal={Biometrika},
  volume={30},
  number={1-2},
  pages={81--93},
  year={1938},
  publisher={Oxford University Press}
}

@article{ca468a70-0be4-389a-b0b9-5dd1ff52b33f,
 ISSN = {00029556},
 URL = {http://www.jstor.org/stable/1412159},
 author = {C. Spearman},
 journal = {The American Journal of Psychology},
 number = {1},
 pages = {72--101},
 publisher = {University of Illinois Press},
 title = {The Proof and Measurement of Association between Two Things},
 urldate = {2025-10-16},
 volume = {15},
 year = {1904}
}

@article{lawrence1989concordance,
  title={A concordance correlation coefficient to evaluate reproducibility},
  author={Lawrence, I and Lin, Kuei},
  journal={Biometrics},
  pages={255--268},
  year={1989},
  publisher={JSTOR}
}

@article{cronbach1951coefficient,
  title={Coefficient alpha and the internal structure of tests},
  author={Cronbach, Lee J},
  journal={psychometrika},
  volume={16},
  number={3},
  pages={297--334},
  year={1951},
  publisher={Springer-Verlag}
}

@book{altman1990practical,
  title={Practical statistics for medical research},
  author={Altman, Douglas G},
  year={1990},
  publisher={Chapman and Hall/CRC}
}

@inproceedings{fournier2013evaluating,
  title={Evaluating text segmentation using boundary edit distance},
  author={Fournier, Chris},
  booktitle={Proceedings of the 51st Annual Meeting of the Association for Computational Linguistics (Volume 1: Long Papers)},
  pages={1702--1712},
  year={2013}
}

@article{pevzner2002critique,
  title={A critique and improvement of an evaluation metric for text segmentation},
  author={Pevzner, Lev and Hearst, Marti A},
  journal={Computational Linguistics},
  volume={28},
  number={1},
  pages={19--36},
  year={2002},
  publisher={MIT Press One Rogers Street, Cambridge, MA 02142-1209, USA journals-info~…}
}

@article{beeferman1999statistical,
  title={Statistical models for text segmentation},
  author={Beeferman, Doug and Berger, Adam and Lafferty, John},
  journal={Machine learning},
  volume={34},
  number={1},
  pages={177--210},
  year={1999},
  publisher={Springer}
}

@article{chhun-etal-2024-language,
    title = "Do Language Models Enjoy Their Own Stories? Prompting Large Language Models for Automatic Story Evaluation",
    author = "Chhun, Cyril  and
      Suchanek, Fabian M.  and
      Clavel, Chlo{\'e}",
    journal = "Transactions of the Association for Computational Linguistics",
    volume = "12",
    year = "2024",
    address = "Cambridge, MA",
    publisher = "MIT Press",
    url = "https://aclanthology.org/2024.tacl-1.62/",
    doi = "10.1162/tacl_a_00689",
    pages = "1122--1142"
}

@article{gwet2001handbook,
  title={Handbook of inter-rater reliability},
  author={Gwet, Kilem},
  journal={Gaithersburg, MD: STATAXIS Publishing Company},
  pages={223--246},
  year={2001}
}

@inproceedings{mohammad-etal-2018-semeval,
    title = "{S}em{E}val-2018 Task 1: Affect in Tweets",
    author = "Mohammad, Saif  and
      Bravo-Marquez, Felipe  and
      Salameh, Mohammad  and
      Kiritchenko, Svetlana",
    editor = "Apidianaki, Marianna  and
      Mohammad, Saif M.  and
      May, Jonathan  and
      Shutova, Ekaterina  and
      Bethard, Steven  and
      Carpuat, Marine",
    booktitle = "Proceedings of the 12th International Workshop on Semantic Evaluation",
    month = jun,
    year = "2018",
    address = "New Orleans, Louisiana",
    publisher = "Association for Computational Linguistics",
    url = "https://aclanthology.org/S18-1001/",
    doi = "10.18653/v1/S18-1001",
    pages = "1--17"
}

@article{bennett,
    author = {BENNETT, E. M. and ALPERT, R. and GOLDSTEIN, A. C.},
    title = {Communications Through Limited-Response Questioning*},
    journal = {Public Opinion Quarterly},
    volume = {18},
    number = {3},
    pages = {303-308},
    year = {1954},
    month = {01},
    issn = {0033-362X},
    doi = {10.1086/266520},
    url = {https://doi.org/10.1086/266520},
    eprint = {https://academic.oup.com/poq/article-pdf/18/3/303/5384778/18-3-303.pdf},
}

@article{fleisig2023majority,
  title={When the majority is wrong: Modeling annotator disagreement for subjective tasks},
  author={Fleisig, Eve and Abebe, Rediet and Klein, Dan},
  journal={arXiv preprint arXiv:2305.06626},
  year={2023}
}

@article{yang2019predicting,
  title={Predicting annotation difficulty to improve task routing and model performance for biomedical information extraction},
  author={Yang, Yinfei and Agarwal, Oshin and Tar, Chris and Wallace, Byron C and Nenkova, Ani},
  journal={arXiv preprint arXiv:1905.07791},
  year={2019}
}

@article{wang2013perspectives,
  title={Perspectives on crowdsourcing annotations for natural language processing},
  author={Wang, Aobo and Hoang, Cong Duy Vu and Kan, Min-Yen},
  journal={Language resources and evaluation},
  volume={47},
  number={1},
  pages={9--31},
  year={2013},
  publisher={Springer}
}

@inproceedings{mehta-srikumar-2023-verifying,
    title = "Verifying Annotation Agreement without Multiple Experts: A Case Study with {G}ujarati {SNACS}",
    author = "Mehta, Maitrey  and
      Srikumar, Vivek",
    editor = "Rogers, Anna  and
      Boyd-Graber, Jordan  and
      Okazaki, Naoaki",
    booktitle = "Findings of the Association for Computational Linguistics: ACL 2023",
    month = jul,
    year = "2023",
    address = "Toronto, Canada",
    publisher = "Association for Computational Linguistics",
    url = "https://aclanthology.org/2023.findings-acl.696/",
    doi = "10.18653/v1/2023.findings-acl.696",
    pages = "10941--10958"
}

@article{huang2020coda,
  title={Coda-19: Using a non-expert crowd to annotate research aspects on 10,000+ abstracts in the covid-19 open research dataset},
  author={Huang, Ting-Hao'Kenneth' and Huang, Chieh-Yang and Ding, Chien-Kuang Cornelia and Hsu, Yen-Chia and Giles, C Lee},
  journal={arXiv preprint arXiv:2005.02367},
  year={2020}
}

@inproceedings{zlabinger2020dexa,
  title={DEXA: supporting non-expert annotators with dynamic examples from experts},
  author={Zlabinger, Markus and Sabou, Marta and Hofst{\"a}tter, Sebastian and Sertkan, Mete and Hanbury, Allan},
  booktitle={Proceedings of the 43rd International ACM SIGIR Conference on Research and Development in Information Retrieval},
  pages={2109--2112},
  year={2020}
}

@article{yang2023assessing,
  title={Assessing inter-annotator agreement for medical image segmentation},
  author={Yang, Feng and Zamzmi, Ghada and Angara, Sandeep and Rajaraman, Sivaramakrishnan and Aquilina, Andr{\'e} and Xue, Zhiyun and Jaeger, Stefan and Papagiannakis, Emmanouil and Antani, Sameer K},
  journal={IEEE Access},
  volume={11},
  pages={21300--21312},
  year={2023},
  publisher={IEEE}
}

@article{schlicher2021flexible,
  title={Flexible, self-determined… and unhealthy? An empirical study on somatic health among crowdworkers},
  author={Schlicher, Katharina D and Schulte, Julian and Reimann, Mareike and Maier, G{\"u}nter W},
  journal={Frontiers in Psychology},
  volume={12},
  pages={724966},
  year={2021},
  publisher={Frontiers Media SA}
}

@inproceedings{wei2018clinical,
  title={Clinical text annotation--what factors are associated with the cost of time?},
  author={Wei, Qiang and Franklin, Amy and Cohen, Trevor and Xu, Hua},
  booktitle={AMIA Annual Symposium Proceedings},
  volume={2018},
  pages={1552},
  year={2018}
}

@article{lim2024towards,
  title={Towards Fair Pay and Equal Work: Imposing View Time Limits in Crowdsourced Image Classification},
  author={Lim, Gordon and Larson, Stefan and Huang, Yu and Leach, Kevin},
  journal={arXiv preprint arXiv:2412.00260},
  year={2024}
}

@article{wu2022time,
  title={Time pressure changes how people explore and respond to uncertainty},
  author={Wu, Charley M and Schulz, Eric and Pleskac, Timothy J and Speekenbrink, Maarten},
  journal={Scientific reports},
  volume={12},
  number={1},
  pages={4122},
  year={2022},
  publisher={Nature Publishing Group UK London}
}

@inproceedings{shmueli-etal-2021-beyond,
    title = "Beyond Fair Pay: Ethical Implications of {NLP} Crowdsourcing",
    author = "Shmueli, Boaz  and
      Fell, Jan  and
      Ray, Soumya  and
      Ku, Lun-Wei",
    editor = "Toutanova, Kristina  and
      Rumshisky, Anna  and
      Zettlemoyer, Luke  and
      Hakkani-Tur, Dilek  and
      Beltagy, Iz  and
      Bethard, Steven  and
      Cotterell, Ryan  and
      Chakraborty, Tanmoy  and
      Zhou, Yichao",
    booktitle = "Proceedings of the 2021 Conference of the North American Chapter of the Association for Computational Linguistics: Human Language Technologies",
    month = jun,
    year = "2021",
    address = "Online",
    publisher = "Association for Computational Linguistics",
    url = "https://aclanthology.org/2021.naacl-main.295/",
    doi = "10.18653/v1/2021.naacl-main.295",
    pages = "3758--3769"
}

@inproceedings{rogstadius2011assessment,
  title={An assessment of intrinsic and extrinsic motivation on task performance in crowdsourcing markets},
  author={Rogstadius, Jakob and Kostakos, Vassilis and Kittur, Aniket and Smus, Boris and Laredo, Jim and Vukovic, Maja},
  booktitle={Proceedings of the international AAAI conference on web and social media},
  volume={5},
  number={1},
  pages={321--328},
  year={2011}
}

@article{salminen2023fair,
  title={Fair compensation of crowdsourcing work: the problem of flat rates},
  author={Salminen, Joni and Kamel, Ahmed Mohamed Sayed and Jung, Soon-Gyo and Mustak, Mekhail and Jansen, Bernard J},
  journal={Behaviour \& Information Technology},
  volume={42},
  number={16},
  pages={2871--2892},
  year={2023},
  publisher={Taylor \& Francis}
}

@inproceedings{snow-etal-2008-cheap,
    title = "Cheap and Fast {--} But is it Good? Evaluating Non-Expert Annotations for Natural Language Tasks",
    author = "Snow, Rion  and
      O{'}Connor, Brendan  and
      Jurafsky, Daniel  and
      Ng, Andrew",
    editor = "Lapata, Mirella  and
      Ng, Hwee Tou",
    booktitle = "Proceedings of the 2008 Conference on Empirical Methods in Natural Language Processing",
    month = oct,
    year = "2008",
    address = "Honolulu, Hawaii",
    publisher = "Association for Computational Linguistics",
    url = "https://aclanthology.org/D08-1027/",
    pages = "254--263"
}

@inproceedings{wan2023everyone,
  title={Everyone’s voice matters: Quantifying annotation disagreement using demographic information},
  author={Wan, Ruyuan and Kim, Jaehyung and Kang, Dongyeop},
  booktitle={Proceedings of the AAAI Conference on Artificial Intelligence},
  volume={37},
  number={12},
  pages={14523--14530},
  year={2023}
}

@inproceedings{sandri-etal-2023-dont,
    title = "Why Don{'}t You Do It Right? Analysing Annotators' Disagreement in Subjective Tasks",
    author = "Sandri, Marta  and
      Leonardelli, Elisa  and
      Tonelli, Sara  and
      Jezek, Elisabetta",
    editor = "Vlachos, Andreas  and
      Augenstein, Isabelle",
    booktitle = "Proceedings of the 17th Conference of the European Chapter of the Association for Computational Linguistics",
    month = may,
    year = "2023",
    address = "Dubrovnik, Croatia",
    publisher = "Association for Computational Linguistics",
    url = "https://aclanthology.org/2023.eacl-main.178/",
    doi = "10.18653/v1/2023.eacl-main.178",
    pages = "2428--2441"
}

@article{rodriguez-barroso-etal-2024-federated,
    title = "Federated Learning for Exploiting Annotators' Disagreements in Natural Language Processing",
    author = "Rodr{\'i}guez-Barroso, Nuria  and
      C{\'a}mara, Eugenio Mart{\'i}nez  and
      Collados, Jose Camacho  and
      Luz{\'o}n, M. Victoria  and
      Herrera, Francisco",
    journal = "Transactions of the Association for Computational Linguistics",
    volume = "12",
    year = "2024",
    address = "Cambridge, MA",
    publisher = "MIT Press",
    url = "https://aclanthology.org/2024.tacl-1.35/",
    doi = "10.1162/tacl_a_00664",
    pages = "630--648"
}

@inproceedings{xu-etal-2024-leveraging,
    title = "Leveraging Annotator Disagreement for Text Classification",
    author = {Xu, Jin  and
      Theune, Mari{\"e}t  and
      Braun, Daniel},
    editor = "Abbas, Mourad  and
      Freihat, Abed Alhakim",
    booktitle = "Proceedings of the 7th International Conference on Natural Language and Speech Processing (ICNLSP 2024)",
    month = oct,
    year = "2024",
    address = "Trento",
    publisher = "Association for Computational Linguistics",
    url = "https://aclanthology.org/2024.icnlsp-1.1/",
    pages = "1--10"
}

@article{reidsma2008reliability,
  title={Reliability measurement without limits},
  author={Reidsma, Dennis and Carletta, Jean},
  journal={Computational Linguistics},
  volume={34},
  number={3},
  pages={319--326},
  year={2008},
  publisher={MIT Press One Rogers Street, Cambridge, MA 02142-1209, USA journals-info~…}
}

@article{Cohen1960,
  title={A Coefficient of Agreement for Nominal Scales},
  author={Cohen, Jacob},
  journal={Educational and Psychological Measurement},
  volume={20},
  number={1},
  pages={37--46},
  year={1960},
  publisher={SAGE Publications}
}

@article{Fleiss1971,
  title={Measuring Nominal Scale Agreement among Many Raters},
  author={Fleiss, Joseph L.},
  journal={Psychological Bulletin},
  volume={76},
  number={5},
  pages={378--382},
  year={1971},
  publisher={American Psychological Association}
}

@book{Krippendorff2013,
  title={Content Analysis: An Introduction to Its Methodology},
  author={Krippendorff, Klaus},
  edition={3rd},
  year={2013},
  publisher={SAGE Publications}
}

@inproceedings{Pevzner2002,
  title={A Critique and Improvement of the P$_k$ Metric for Text Segmentation},
  author={Pevzner, Leonid and Hearst, Marti},
  booktitle={Proceedings of the 40th Annual Meeting on Association for Computational Linguistics (ACL)},
  pages={27--36},
  year={2002},
  organization={Association for Computational Linguistics}
}

@article{hallgren2012computing,
  title={Computing inter-rater reliability for observational data: an overview and tutorial},
  author={Hallgren, Kevin A},
  journal={Tutorials in quantitative methods for psychology},
  volume={8},
  number={1},
  pages={23},
  year={2012}
}

@article{koo2016guideline,
  title={A guideline of selecting and reporting intraclass correlation coefficients for reliability research},
  author={Koo, Terry K and Li, Mae Y},
  journal={Journal of chiropractic medicine},
  volume={15},
  number={2},
  pages={155--163},
  year={2016},
  publisher={Elsevier}
}

@article{kohen1960coefficient,
  title={A coefficient of agreement for nominal scale},
  author={Kohen, Jacob},
  journal={Educ Psychol Meas},
  volume={20},
  pages={37--46},
  year={1960}
}

@article{liu2024aligning,
  title={Aligning with human judgement: The role of pairwise preference in large language model evaluators},
  author={Liu, Yinhong and Zhou, Han and Guo, Zhijiang and Shareghi, Ehsan and Vuli{\'c}, Ivan and Korhonen, Anna and Collier, Nigel},
  journal={arXiv preprint arXiv:2403.16950},
  year={2024}
}

@inproceedings{deng-etal-2023-annotate,
    title = "You Are What You Annotate: Towards Better Models through Annotator Representations",
    author = "Deng, Naihao  and
      Zhang, Xinliang  and
      Liu, Siyang  and
      Wu, Winston  and
      Wang, Lu  and
      Mihalcea, Rada",
    editor = "Bouamor, Houda  and
      Pino, Juan  and
      Bali, Kalika",
    booktitle = "Findings of the Association for Computational Linguistics: EMNLP 2023",
    month = dec,
    year = "2023",
    address = "Singapore",
    publisher = "Association for Computational Linguistics",
    url = "https://aclanthology.org/2023.findings-emnlp.832/",
    doi = "10.18653/v1/2023.findings-emnlp.832",
    pages = "12475--12498"
}

@article{tam2024framework,
  title={A framework for human evaluation of large language models in healthcare derived from literature review},
  author={Tam, Thomas Yu Chow and Sivarajkumar, Sonish and Kapoor, Sumit and Stolyar, Alisa V and Polanska, Katelyn and McCarthy, Karleigh R and Osterhoudt, Hunter and Wu, Xizhi and Visweswaran, Shyam and Fu, Sunyang and others},
  journal={NPJ digital medicine},
  volume={7},
  number={1},
  pages={258},
  year={2024},
  publisher={Nature Publishing Group UK London}
}

@proceedings{humeval-2024-human,
    title = "Proceedings of the Fourth Workshop on Human Evaluation of NLP Systems (HumEval) @ LREC-COLING 2024",
    editor = "Balloccu, Simone  and
      Belz, Anya  and
      Huidrom, Rudali  and
      Reiter, Ehud  and
      Sedoc, Joao  and
      Thomson, Craig",
    month = may,
    year = "2024",
    address = "Torino, Italia",
    publisher = "ELRA and ICCL",
    url = "https://aclanthology.org/2024.humeval-1.0/"
}

@inproceedings{bavaresco-etal-2025-llms,
    title = "{LLM}s instead of Human Judges? A Large Scale Empirical Study across 20 {NLP} Evaluation Tasks",
    author = "Bavaresco, Anna  and
      Bernardi, Raffaella  and
      Bertolazzi, Leonardo  and
      Elliott, Desmond  and
      Fern{\'a}ndez, Raquel  and
      Gatt, Albert  and
      Ghaleb, Esam  and
      Giulianelli, Mario  and
      Hanna, Michael  and
      Koller, Alexander  and
      Martins, Andre  and
      Mondorf, Philipp  and
      Neplenbroek, Vera  and
      Pezzelle, Sandro  and
      Plank, Barbara  and
      Schlangen, David  and
      Suglia, Alessandro  and
      Surikuchi, Aditya K  and
      Takmaz, Ece  and
      Testoni, Alberto",
    editor = "Che, Wanxiang  and
      Nabende, Joyce  and
      Shutova, Ekaterina  and
      Pilehvar, Mohammad Taher",
    booktitle = "Proceedings of the 63rd Annual Meeting of the Association for Computational Linguistics (Volume 2: Short Papers)",
    month = jul,
    year = "2025",
    address = "Vienna, Austria",
    publisher = "Association for Computational Linguistics",
    url = "https://aclanthology.org/2025.acl-short.20/",
    doi = "10.18653/v1/2025.acl-short.20",
    pages = "238--255",
    ISBN = "979-8-89176-252-7"
}

@inproceedings{chhun-etal-2022-human,
    title = "Of Human Criteria and Automatic Metrics: A Benchmark of the Evaluation of Story Generation",
    author = "Chhun, Cyril  and
      Colombo, Pierre  and
      Suchanek, Fabian M.  and
      Clavel, Chlo{\'e}",
    editor = "Calzolari, Nicoletta  and
      Huang, Chu-Ren  and
      Kim, Hansaem  and
      Pustejovsky, James  and
      Wanner, Leo  and
      Choi, Key-Sun  and
      Ryu, Pum-Mo  and
      Chen, Hsin-Hsi  and
      Donatelli, Lucia  and
      Ji, Heng  and
      Kurohashi, Sadao  and
      Paggio, Patrizia  and
      Xue, Nianwen  and
      Kim, Seokhwan  and
      Hahm, Younggyun  and
      He, Zhong  and
      Lee, Tony Kyungil  and
      Santus, Enrico  and
      Bond, Francis  and
      Na, Seung-Hoon",
    booktitle = "Proceedings of the 29th International Conference on Computational Linguistics",
    month = oct,
    year = "2022",
    address = "Gyeongju, Republic of Korea",
    publisher = "International Committee on Computational Linguistics",
    url = "https://aclanthology.org/2022.coling-1.509/",
    pages = "5794--5836"
}

@inproceedings{ghosh-etal-2024-compare,
    title = "Compare without Despair: Reliable Preference Evaluation with Generation Separability",
    author = "Ghosh, Sayan  and
      Srinivasan, Tejas  and
      Swayamdipta, Swabha",
    editor = "Al-Onaizan, Yaser  and
      Bansal, Mohit  and
      Chen, Yun-Nung",
    booktitle = "Findings of the Association for Computational Linguistics: EMNLP 2024",
    month = nov,
    year = "2024",
    address = "Miami, Florida, USA",
    publisher = "Association for Computational Linguistics",
    url = "https://aclanthology.org/2024.findings-emnlp.747/",
    doi = "10.18653/v1/2024.findings-emnlp.747",
    pages = "12787--12805"
}

@inproceedings{braylan2022measuring,
  title={Measuring annotator agreement generally across complex structured, multi-object, and free-text annotation tasks},
  author={Braylan, Alexander and Alonso, Omar and Lease, Matthew},
  booktitle={Proceedings of the ACM Web Conference 2022},
  pages={1720--1730},
  year={2022}
}

@article{hayes2007answering,
  title={Answering the call for a standard reliability measure for coding data},
  author={Hayes, Andrew F and Krippendorff, Klaus},
  journal={Communication methods and measures},
  volume={1},
  number={1},
  pages={77--89},
  year={2007},
  publisher={Taylor \& Francis}
}

@article{Mathet2015,
  title={The Unified and Holistic Method Gamma for Inter-Annotator Agreement Measure and Alignment},
  author={Mathet, Yann and Widl{\"o}cher, Antoine and M{\'e}tivier, Jean-Philippe},
  journal={Computational Linguistics},
  volume={41},
  number={3},
  pages={437--479},
  year={2015},
  publisher={MIT Press}
}

@article{McGraw1996,
  title={Forming Inferences About Some Intraclass Correlation Coefficients},
  author={McGraw, Kimberly O. and Wong, Sean P.},
  journal={Psychological Methods},
  volume={1},
  number={1},
  pages={30--46},
  year={1996},
  publisher={American Psychological Association}
}

\end{document}